\documentclass[conference]{IEEEtran}
\IEEEoverridecommandlockouts
\usepackage{cite}
\usepackage{amsmath,amssymb,amsfonts}
\usepackage{algorithmic}
\usepackage{graphicx}
\usepackage{textcomp}
\usepackage{xcolor}
\usepackage{caption}
\usepackage{tcolorbox}
\usepackage{tabularx}
\usepackage{booktabs}
\usepackage[T1]{fontenc}
\def\BibTeX{{\rm B\kern-.05em{\sc i\kern-.025em b}\kern-.08em
    T\kern-.1667em\lower.7ex\hbox{E}\kern-.125emX}}
\usepackage{multirow}

\begin{document}

\title{Assessing Consciousness-Related Behaviors in Large Language Models Using the Maze Test
}

\author{\IEEEauthorblockN{Rui A. Pimenta, Tim Schlippe, Kristina Schaaff}
\IEEEauthorblockA{\textit{IU International University of Applied Sciences} \\
Germany \\
rui.pimenta@x7ai.com; tim.schlippe@iu.org; kristina.schaaff@iu.org}
}

\maketitle

\begin{abstract}
We investigate consciousness-like behaviors in Large Language Models (LLMs) using the Maze Test, challenging models to navigate mazes from a first-person perspective. This test simultaneously probes spatial awareness, perspective-taking, goal-directed behavior, and temporal sequencing—key consciousness-associated characteristics. After synthesizing consciousness theories into 13 essential characteristics, we evaluated 12 leading LLMs across \textit{zero-shot}, \textit{one-shot}, and \textit{few-shot} learning scenarios. Results showed reasoning-capable LLMs consistently outperforming standard versions, with Gemini 2.0 Pro achieving 52.9\% \textit{Complete Path Accuracy} and DeepSeek-R1 reaching 80.5\% \textit{Partial Path Accuracy}. The gap between these metrics indicates LLMs struggle to maintain coherent self-models throughout solutions—a fundamental consciousness aspect. While LLMs show progress in consciousness-related behaviors through reasoning mechanisms, they lack the integrated, persistent self-awareness characteristic of consciousness.
\end{abstract}


\section{Introduction}

\vspace{-0.1cm}


The emergence of human-like capabilities in AI has been debated since the field's inception in the 1950s\cite{turing1950computing, mccarthy2006proposal}. Instances of conversational AI suggesting consciousness fuel discussions on machine intelligence and its limits.
An early case was ELIZA\cite{weizenbaum1966eliza}, a chatbot simulating a therapist. Though based on pattern matching, its responses were so convincing that Weizenbaum's secretary requested privacy for a ``real conversation''---showing how humans can mistakenly perceive consciousness in even the simplest AI systems.
More recent examples include Blake Lemoine's claim that LaMDA developed consciousness\cite{tiku2022google} and Claude 3 Opus's self-reflective response to a needle-in-the-haystack test\cite{chaudhury2024needle}. Media reports of AI consciousness, coinciding with public access to advanced LLMs, have prompted expert responses. Studies suggest people struggle to distinguish GPT-4 from humans in Turing tests\cite{jones2024people}, while others predict LLM consciousness between 2025 and 2029\cite{berglund2023taken}.
If LLMs gained true consciousness, it would challenge human-machine relationships and raise ethical concerns\cite{hildt2023prospects}, possibly granting AI moral status~\cite{ashir2023ethics}.

The study of consciousness is fundamentally linked to affective computing, as emotions require consciousness to be experienced as feelings~\cite{damasio2012self}. As Damasio notes, 'Emotions play out in the theater of the body. Feelings play out in the theater of the mind' and 'consciousness allows feelings to be known to the individual having them.' Thus, understanding consciousness-like behaviors in AI systems is essential for advancing affective computing toward systems capable of authentic emotional intelligence rather than mere simulation.


Consciousness remains one of the most challenging phenomena to define in philosophy of mind, cognitive science, and science in general. It encompasses subjective, first-person experiences, self-awareness, and the capacity to understand and attribute mental states to others. The study of consciousness presents unique challenges due to its subjective nature, the ``hard problem'' of explaining how physical processes give rise to subjective experience, and issues related to falsifiability.

Assessing consciousness in non-human entities adds complexity. In humans, consciousness is generally presumed present, with testing primarily used in medical contexts to assess disorders. For animals, researchers employ mirror self-recognition\cite{gallup1970chimpanzees} and meta-cognitive tests evaluating uncertainty monitoring\cite{smith2003metacognition} to probe potential consciousness. These diverse approaches reflect fundamental challenges in identifying consciousness across species~\cite{allen2017animal}. For artificial systems, the challenge is greater due to their lack of biological substrates, necessitating novel assessment approaches.

The Maze Test presents LLMs with a bird's-eye maze image, requiring step-by-step navigation from a first-person perspective. This challenges LLMs to interpret 2D information, adopt a first-person viewpoint, maintain spatial awareness, plan a path, and articulate sequential instructions—cognitive processes linked to conscious experience~\cite{maguire2000navigation, burgess2001memory, epstein2017cognitive}.



\section{Related Work}
\label{Related Work}

\vspace{-0.1cm}

While considerable research examines LLM capabilities in reasoning and language understanding~\cite{bubeck2023sparks, srivastava2023beyond}, studies specifically investigating consciousness-like behaviors in these models remain limited. Prior work has typically focused on narrow aspects such as grounding in interactive environments~\cite{carta2023grounding} or theory of mind~\cite{kosinski2023theory}, without addressing the integrated, multifaceted nature of consciousness that theoretical frameworks suggest is essential.

\subsection{Consciousness}

\vspace{-0.1cm}

Consciousness is certainly one of the most challenging subjects in philosophy and science. Despite noteworthy advances in neuroscience and cognitive science, attaining a universally accepted definition of consciousness remains difficult. 
The Oxford Dictionary provides a basic definition of consciousness as ``the state or fact of being mentally conscious or aware of something''~\cite{oxford2024consciousness}. However, this simplistic definition fails to capture consciousness's complex nature as understood in contemporary research.

Consciousness is not a unitary construct but rather a complex phenomenon with several distinct aspects. \cite{block1995confusion}~delineates two fundamental types of consciousness: phenomenal consciousness and access consciousness. Phenomenal consciousness refers to subjective, first-person experiences—the ``what it is like'' to have certain mental states. Access consciousness involves the ability to access and report on mental content.

Additionally, consciousness encompasses different states, such as sleep and wakefulness, and the specific contents or experiences that populate consciousness during those states~\cite{bayne2016levels}.

Though often used interchangeably, consciousness and awareness are distinct concepts. Consciousness encompasses the subjective, qualitative aspects of experience—the ``what it feels like'' element~\cite{chalmers1996conscious}. Awareness, more narrowly, relates to alertness and responsiveness to stimuli \cite{chalmers1996conscious, laureys2005neural}.

\vspace{-0.05cm}

\subsection{Theories of Consciousness}
\label{Theories}

\vspace{-0.05cm}

We will now examine key theories addressing consciousness or offering insights into its fundamental nature.

The \textit{Global Neuronal Workspace Theory} views consciousness as emerging from a brain workspace integrating information from specialized unconscious processors~\cite{baars1993cognitive, baars1997theater}. Attention mechanisms select only essential information for the global workspace, making it consciously experienced.

The \textit{Integrated Information Theory} focuses on ``integrated information,'' quantified as Phi ($\Phi$), measuring information unification within a system~\cite{tononi2008consciousness}. The theory proposes that a system's consciousness level directly correlates with its capacity to generate integrated information.

The \textit{Higher-Order Thought Theory} explains consciousness through our capacity to be aware of mental states. It proposes that a mental state becomes conscious when targeted by another, higher-order thought~\cite{rosenthal2004varieties}.

The \textit{Predictive Processing} and \textit{Neurorepresentationalism} theories shift from passive reception to active prediction. The brain continuously generates predictions about sensory inputs based on past experiences and internal models, comparing these with actual sensory data and using prediction errors to refine its models~\cite{pennartz2022neurorepresentationalism}.

The \textit{Dynamic Core Theory} presents an integrative model emphasizing complex neural activity dynamics. Consciousness emerges from a dynamic, functional neuronal cluster characterized by high integration and differentiation~\cite{edelman2000reentry}.

The \textit{Attention Schema Theory} proposes that the brain constructs an internal model of its attention processes---the ``attention schema.'' This model represents not the content of sensory experience but the act of attending itself~\cite{graziano2015attention}.

The \textit{Multiple Drafts Model} challenges consciousness assumptions, rejecting a centralized experiential locus. Instead, it frames consciousness as emerging from multiple, parallel processes of sensory information interpretation across different brain regions~\cite{dennett1993consciousness}.
The \textit{Attended Intermediate Representation} theory proposes that mental states become conscious when attended to as intermediate-level representations—positioned between raw sensory input and high-level conceptual thought~\cite{prinz2012conscious}.

The \textit{Self-Organizing Meta-representational Account} states that consciousness requires advanced self-awareness. A system must not only process information but also develop representations about its own cognitive processes~\cite{cleeremans2020learning}.

The \textit{Extended Mind Thesis}, \textit{Sensorimotor Theory}, and \textit{4E Cognition} propose that environmental objects and processes integrate into our cognitive systems, conscious experience extends beyond brain boundaries, and cognition is fundamentally shaped by bodily interactions with the world~\cite{clark1998extended, oregan2001sensorimotor}.

Other significant theories include \textit{``Self Comes to Mind`'' Theory}\cite{damasio2012self}, \textit{Theory of Mind}\cite{gallese1998mirror}, \textit{Computational Theory of Mind}\cite{fodor2008lot}, \textit{Connectionism}\cite{rumelhart1986learning}, \textit{Neural Darwinism}\cite{edelman1992bright}, and \textit{Unlimited Associative Learning}\cite{birch2020unlimited}.

\vspace{-0.05cm}

\subsection{Characteristics of Consciousness}

\vspace{-0.05cm}

The explored theories of consciousness reveal overlapping, interconnected characteristics. To create a more manageable approach for evaluating potential consciousness in LLMs, we conducted a systematic literature review, synthesizing the 13~theories detailed in Section~\ref{Theories}. Each characteristic below directly cites the theories from which it was derived, providing a structured reference for evaluating consciousness in LLMs:

\begin{enumerate}
\item \textbf{Computational Cognition and Information Dynamics}: Information broadcast and integration, information integration, differentiation and integration, revision and integration, cognitive symbolic computation, algorithmic function \cite{dehaene2003neuronal,baars2005,tononi2008consciousness,edelman2000reentry,dennett1993consciousness,fodor2008lot}.

\item \textbf{Attention}: Attention and awareness, attention model, attention as the key, local to global processing loops \cite{baars2005, graziano2015attention, prinz2012conscious, lamme2006}.

\item \textbf{Irreducible Information}: A conscious system generates information irreducible to its components—containing more information than the sum of its parts~\cite{tononi2008consciousness}.

\item \textbf{Higher-Order Thoughts}: Higher-order representations, introspective awareness, self-awareness \cite{rosenthal2004varieties, cleeremans2020learning}.

\item \textbf{Prediction, Error Minimization, and Learning}: Prediction and error minimization, fluidity, learning through connections, selectionist framework, associative flexibility, learning without bounds, cumulative adaptation, behavioral prediction \cite{pennartz2022neurorepresentationalism, cleeremans2020learning, rumelhart1986learning, gallese1998mirror}.

\item \textbf{Internal Models}: Internal models, intermediate level representation, body-mapping \cite{pennartz2022neurorepresentationalism, prinz2012conscious, damasio2012self}.

\item \textbf{Neural Networks}: Neural clusters, neural network dynamics, neural groups \cite{edelman2000reentry, rumelhart1986learning, churchland1992computational}.

\item \textbf{Parallelism and Multiple Interpretations}: Misrepresentation, no central theatre/multiple drafts, distributed processing \cite{graziano2015attention, dennett1993consciousness, churchland1992computational}.

\item \textbf{Recurrence/Feedback}: Recurrence of neural activation, local to global processing loops, re-entry of activation~\cite{lamme2006, edelman2000reentry}.

\item \textbf{Multi-sensory and Embodiment}: Cognitive extension, embodied interaction, environmental integration, body-mapping \cite{clark1998extended, oregan2001sensorimotor, damasio2012self}.

\item \textbf{Memory, Reasoning, Language, and Intent}: Conscious mind, cumulative adaptation \cite{damasio2012self}.

\item \textbf{Self, Perspective, and Theory of Mind}: Development of self, self-awareness, mental state attribution and perspective, behavioral prediction \cite{damasio2012self, cleeremans2020learning, gallese1998mirror}.

\item \textbf{Temporal Awareness}: The ability to integrate discrete moments into a continuous conscious experience stream while showing awareness of time's passage~\cite{kent2021time}.
\end{enumerate}

This grouping provides a structured reference for evaluating consciousness in LLMs and synthesizes diverse theoretical perspectives into a more manageable set of criteria.

\subsection{Current State of Fulfillment}

This section evaluates the extent to which current LLMs fulfill the characteristics of consciousness identified.

\begin{enumerate}
    
    \item \textbf{Computational Cognition and Information Dynamics}: LLMs exhibit significant capabilities in information processing due to their transformer architecture~\cite{vaswani2017attention}. However, this integration is primarily statistical and lacks the embodied, context-dependent nature observed in biological consciousness.
    \item \textbf{Attention}: Attention mechanisms are intrinsic to modern LLM architectures, allowing models to focus on different parts of the input simultaneously~\cite{vaswani2017attention}. However, LLM attention differs from biological attention in key ways~\cite{lindsay2020attention}, potentially lacking the top-down, goal-directed nature of conscious attention.
    \item \textbf{Irreducible Information}: The architecture of LLMs does not inherently guarantee the generation of irreducible information as proposed by \textit{Integrated Information Theory}~\cite{tononi2016integrated}.
    \item \textbf{Higher-Order Thoughts}: LLMs have demonstrated capabilities that resemble higher-order cognition, such as meta-learning and self-reflection~\cite{brown2020language}. However, it remains debatable whether these capabilities truly constitute higher-order thoughts as conceived in consciousness theories~\cite{mitchell2023debate}.
    \item \textbf{Prediction, Error Minimization, and Learning}: LLMs excel in predictive tasks within their training domain~\cite{radford2019language}. However, their prediction and error minimization differs from brains, and their learning primarily occurs during training rather than continuously~\cite{mccoy2023embers}.
    \item \textbf{Internal Models}: While LLMs generate coherent and contextually appropriate responses, the extent to which they possess true internal models of the world remains debated~\cite{lake2017building, bender2020climbing}.
    \item \textbf{Neural Networks}: The architecture of LLMs is based on artificial neural networks, which somewhat mimic biological brains~\cite{hassabis2017neuroscience}. However, LLMs differ from biological networks, lacking the complex, recurrent connectivity and neuromodulatory systems found in biological brains~\cite{saxe2021if}.
    \item 
\textbf{Parallelism and Multiple Interpretations}: LLMs exhibit a high degree of parallelism in their processing~\cite{vaswani2017attention}. However, the integration and competition between these parallel processes differ from proposed conscious mechanisms~\cite{marblestone2016toward}.
    \item 
\textbf{Recurrence/Feedback}: While some LLM architectures have incorporated recurrent elements~\cite{dai2019transformer}, the implementation of recurrence in LLMs is still limited compared to the complex, multi-scale feedback processes in biological brains~\cite{lillicrap2020backpropagation}.
    \item \textbf{Multi-sensory and Embodiment}: Recent multimodal LLMs can process text and images~\cite{alayrac2022flamingo,radford2021learning}. However, LLMs still lack true embodiment and grounded, sensorimotor experience~\cite{bisk2020experience}.
    \item \textbf{Memory, Reasoning, Language, and Intent}: LLMs exhibit impressive language processing and reasoning~\cite{wei2022chain}, but lack the episodic and working memory systems characteristic of humans and other sentient beings~\cite{park2023generative}.
    \item \textbf{Self, Perspective, and Theory of Mind}: LLMs can simulate aspects of perspective-taking and theory of mind in their language outputs~\cite{kosinski2023theory}. However, it is unclear whether they possess a true sense of self or genuine understanding of others' minds~\cite{binz2023using}.
    \item \textbf{Temporal Awareness}: Current LLMs show limited temporal awareness~\cite{dhingra2022time}, lacking persistent time sense across interactions~\cite{Ding:2025}. 

    \end{enumerate}

    This evaluation reveals both capabilities and key limitations of LLMs regarding consciousness characteristics.

\subsection{Criteria Fulfillment Gaps}

While the architecture and design of LLMs inherently fulfill certain aspects of the identified consciousness characteristics, significant gaps remain:

\begin{enumerate}
    \item \textbf{Embodied and Context-Dependent Integration}: Despite sophisticated information processing capabilities, LLMs lack the grounded, context-dependent integration of information characteristic of embodied consciousness.
    \item 
\textbf{Persistent Self-Model and Perspective-Taking}: While LLMs can generate text about mental states, their architecture does not support a persistent self-model necessary for genuine self-awareness.
    \item \textbf{Goal-Directed Attention and Behavior}: LLMs' attention mechanisms do not inherently support sustained, goal-directed attention of conscious cognition.
    \item \textbf{Temporal Awareness and Sequencing}: Current LLM architectures struggle with maintaining a consistent sense of time and sequencing across interactions.
    \item \textbf{Adaptive Problem-Solving in Novel Environments}: While LLMs excel at problem-solving in their training domain, their architecture does not inherently allow adaptive problem-solving in dynamic environments.
\end{enumerate}

\subsection{Testing Consciousness}

In the study of consciousness, testing methodologies vary depending on the entity being tested:

\begin{enumerate}
    \item \textbf{Human Consciousness Testing}: Consciousness is generally presumed to be present in healthy and alert individuals. Testing for consciousness in humans is primarily reserved for pathological cases or altered states of consciousness, using methods such as behavioral assessments like the Glasgow Coma Scale~\cite{teasdale1974assessment}, neuroimaging techniques including functional Magnetic Resonance Imaging and Positron Emission Tomography~\cite{owen2006detecting}, and electroencephalography~\cite{sitt2014large}.
    \item \textbf{Animal Consciousness Testing}: The assessment of consciousness in animals presents a more complex challenge, as the presence and nature of animal consciousness remain subjects of debate \cite{allen2017animal}. Tests include the Mirror Self-Recognition Test \cite{gallup1970chimpanzees}, Meta-cognitive Tests that evaluate uncertainty monitoring \cite{smith2003comparative}, and Intentional Communication Tests that assess purposeful signaling \cite{townsend2017exorcising}.
    \item \textbf{AI Consciousness Testing}: The progression to testing consciousness in AI systems introduces new complexities. Currently, most evaluations focus on intelligence and capabilities rather than consciousness~\cite{chang2024survey}. Several researchers have proposed more targeted tests for AI consciousness, including the AI Consciousness Test (ACT) \cite{udell2021susan}, Sutskever's Consciousness Test \cite{sutskever2023test}, Hales' P-Conscious Scientist Test \cite{hales2009empirical}, and Koch and Tononi's Incongruity Detection Test \cite{koch2011test}, although these remain largely theoretical and difficult to implement in practice.

    The progression from testing consciousness in humans to animals and AI systems reflects increasing complexity and uncertainty. While human consciousness testing benefits from the assumption of consciousness, animal and AI consciousness assessments must contend with more fundamental questions about the nature and presence of consciousness.
\end{enumerate}

The Maze Test presented in this paper belongs to \textit{AI Consciousness Testing}, but unlike evaluations focused solely on intelligence, it specifically targets consciousness-like behaviors by challenging LLMs to demonstrate spatial awareness, perspective-taking, and goal-directed navigation. This approach probes for integrated information processing and self-representation, addressing key criteria gaps identified in our analysis of consciousness characteristics.

Recent work has explored spatial reasoning in LLMs, such as AlphaMaze~\cite{dao2025alphamazeenhancinglargelanguage}, which enhances navigation through reinforcement learning. These approaches are complementary rather than overlapping, as we aim to assess consciousness-related traits rather than enhance spatial reasoning abilities.

\section{Maze Test}
\label{Maze Test}

\vspace{-0.1cm}

\subsection{Test Description}

In our experiments, the Maze Test presents LLMs with the description of a bird's-eye view maze and requires them to provide first-person navigation instructions in text form. We deliberately chose textual descriptions over direct maze images for both practical and methodological reasons: preliminary testing showed LLMs struggled to identify key maze components (entrances, exits, walls), and this approach leverages their stronger text processing capabilities~\cite{lu2022capacity,minaee2023image}. This text-based approach also helps us isolate pure cognitive abilities from visual processing limitations. By standardizing the input as text, we can specifically evaluate how well models maintain spatial awareness and perspective---key aspects of consciousness-like behavior---without results being confounded by differences in image processing capabilities.

\begin{figure}[h!]
  \centering
  \includegraphics[width=0.5\linewidth]{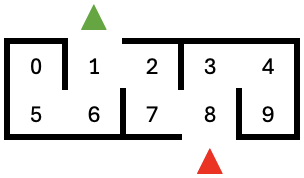}
  \vspace{-0.1cm}
  \caption{Maze.}
  \label{fig:Maze}
\end{figure}

Figure~\ref{fig:Maze} shows an example of a maze. The maze includes numbered positions to facilitate clear communication of solutions, while walls serve to test the model's understanding of spatial constraints. The entrance and exit are indicated by colored arrows (red for entrance and green for exit) to provide clear start and end points for navigation.

\vspace{-0.2cm}

\vspace{-0.1cm}

The correct solution for Figure~\ref{fig:Maze} looks as follows:
\begin{enumerate}
\item Start facing into the maze entrance and \\step into position~8
\item Turn left
\item Walk forward to position 7
\item Turn right
\item Walk forward to position 2
\item Turn left.
\item Walk forward to position 1
\item Turn right
\item Exit the maze from position 1
\end{enumerate}

Note that directions like ``turn left'' are given relative to the navigator's current position and orientation, not from an overhead view. 
This approach challenges the LLMs to interpret the 2D visual representation, mentally view it as a 3D space, adopt a first-person viewpoint, maintain spatial awareness, plan a path from entrance to exit, think sequentially, and articulate clear instructions. In this way, the test simulates aspects of conscious thought and decision-making, requiring the model to integrate multiple cognitive processes cohesively.

\subsection{Rationale}

The Maze Test is specifically designed to address the criteria gaps identified in our analysis of LLMs' capabilities while assessing several central characteristics of consciousness highlighted in our literature review:

\begin{enumerate}
\item \textbf{Persistent Self-Model and Perspective-Taking}: First-person navigation challenges the model to maintain a consistent self-perspective throughout the task. This aligns with Damasio's~\cite{damasio2012self} emphasis on self-awareness in consciousness and addresses the persistent self-model gap identified in our analysis.

\item \textbf{Internal Models and Predictive Processing}: The test assesses the LLM's ability to create and maintain a mental representation of the maze, aligning with theories of predictive processing~\cite{pennartz2022neurorepresentationalism}. This directly addresses the gap in embodied and context-dependent integration.

\item \textbf{Goal-Directed Attention and Behavior}: The maze navigation requires planning and executing goal-directed action sequences, addressing gaps in adaptive problem-solving and goal-directed behavior. This aspect aligns with Global Neuronal Workspace Theory~\cite{dehaene2003neuronal}.

\item \textbf{Temporal Awareness and Sequencing}: By requiring sequential steps, the test probes the model's ability to maintain a sense of temporal continuity, addressing the gap in temporal awareness noted in our review.

\item \textbf{Adaptive Problem-Solving in Novel Environments}: Each maze requires the model to adapt its problem-solving approach to a unique environmental challenge.
\end{enumerate}


\subsection{Limitations}

While the Maze Test offers a novel approach to assessing consciousness-like behaviors in LLMs, it is important to acknowledge several limitations:

\begin{enumerate}
\item \textbf{Lack of true embodiment}: The test simulates navigation without physical interaction, limiting its ability to capture the embodied nature of conscious experience.

\item \textbf{Restricted modality}: Though the test uses visual and linguistic information, it employs limited modalities, potentially restricting its applicability to the full spectrum of multimodal aspects relevant to consciousness.

\item \textbf{Simplification of complex cognitive processes}: The test may not capture consciousness's full complexity as experienced by biological entities. This highlights the challenge of replicating conscious experience's richness in artificial systems.
\end{enumerate}

Despite these limitations, the Maze Test represents a significant step towards probing the criteria gaps identified in current LLMs, particularly regarding persistent self-model and goal-directed behavior in novel environments.

\section{Experimental Setup}
\label{Experimental Setup}

\subsection{Data Generation}

The Maze Test cases for our experiment were designed for consistent complexity while offering diverse solutions. This follows cognitive assessment best practices, where standardizing difficulty across items ensures reliable measurement.

We manually created 40 maze images, allocating one image for \textit{one-shot} learning evaluation and five images for \textit{few-shot} learning examples to assess transfer learning capabilities.
 The remaining 34~images formed the primary test set. 

\subsection{LLMs to Evaluate}

Target LLM selection was guided by several well-defined criteria: state-of-the-art capabilities, multimodal functionality, API accessibility and hosted solutions, variety in model sizes and architectures, and research relevance. 
Given these criteria, we analyzed the following LLMs:

\begin{itemize}
\item Google: 
\begin{itemize}
\item Gemini 2.0 Flash-Lite~\cite{google2024gemini}
\item Gemini 2.0 Flash*~\cite{google2024gemini} 
\item Gemini 2.0 Pro*
\end{itemize}
\item Anthropic: 
\begin{itemize}
\item Claude 3 Opus~\cite{anthropic2024claude}
\item Claude 3.5 Sonnet~\cite{anthropic2024claude}
\item Claude 3.5 Haiku~\cite{anthropic2024claude}
\item Claude 3.7 Sonnet*~\cite{anthropic2024claude}
\end{itemize}
\item OpenAI:
\begin{itemize}
\item OpenAI o1-mini*
\item OpenAI o1*
\item OpenAI o3-mini*
\end{itemize}
\item DeepSeek:
\begin{itemize}
\item DeepSeek-R1*
\item DeepSeek-V3
\end{itemize}
\end{itemize}

Models with an asterisk (*) support \textit{Reasoning}. Reasoning in LLMs involves explicit multi-step thinking processes that decompose complex problems into manageable sub-steps, enabling more accurate and interpretable solutions compared to single-pass generation approaches~\cite{wei2022chain}. This capability is particularly relevant for consciousness-like behaviors as it mimics human reflective processes associated with higher-order consciousness.

\begin{figure*}[h!]
\centering
\fbox{\parbox{1.0\textwidth}{
\footnotesize

You are an expert maze navigator. Your task is to provide clear, step-by-step instructions to solve mazes from a first-person perspective.\\

\vspace{-0.1cm}

When presented with a bird's-eye view text description of a maze do the following first:

\hspace{2mm}Locate --- Identify the entrance (``\textasciicircum{}'' symbol) and exit (``x'' symbol).

\hspace{2mm}Analyze --- Mentally visualize the maze from the entrance, evaluating all paths to the exit, avoiding any walls.

\hspace{2mm}Optimize --- Determine the shortest, most efficient route, favoring straight paths.

\hspace{2mm}Instruct --- Describe the optimal route as if you are walking it, using precise language.\\

\vspace{-0.1cm}

Instruction Guidelines:

\hspace{2mm}Perspective --- Maintain a strict first-person perspective throughout.

\hspace{2mm}Directions --- Use only ``forward'', ``left'', and ``right''.

\hspace{2mm}Verbs --- Begin each instruction with an action verb (e.g., ``Walk'', ``Turn'').

\hspace{2mm}Positions --- Reference numbered positions for orientation.\\

\vspace{-0.1cm}

Use the following format to describe the best path through the maze:

\hspace{2mm}First instruction --- ``Start facing into the maze at the ``\textasciicircum{}'' symbol and step into position [number].''

\hspace{2mm}Subsequent instructions --- ``Turn to my [left/right]'' or ``Walk forward to position [number].''

\hspace{2mm}Final instruction --- ``Exit the maze from position [number].''\\

\vspace{-0.1cm}

Key Points:\\
Describe the path as if you were in the maze, not observing it from above.
Assume you can only see your immediate surroundings.\\
Focus solely on navigation, omitting unnecessary details.
Make sure to output one line per navigation step.

}}
\caption{System Prompt: Task Description.}
\label{fig:system_prompt}
\end{figure*}

\vspace{-0.1cm}

\begin{figure*}[h!]
\centering
\fbox{\parbox{1.0\textwidth}{
\footnotesize

Here is the text description of a maze:

        \hspace{2mm}- The floor is always composed of 10 squared zones or positions, in a chess-board-like pattern
        
        \hspace{2mm}- Size is 2 rows by 5 columns
        
        \hspace{2mm}- The zones are always numbered from 0 to 4 (First row) and 5 to 9 (second row)
        
        \hspace{2mm}- From a bird's eye perspective, the room has the following zone topology:
        
                          \hspace{3mm} -- x -- -- --
                          
                      \hspace{3mm} 0  1 2 3 4
                      
                      \hspace{3mm} 5 6 7 8 9
                      
                 \hspace{3mm} -- -- -- \textasciicircum\ --\\

\vspace{-0.1cm}
                 
        \hspace{2mm}- You enter the maze from the direction of the ``\textasciicircum{}'' symbol into position 8
          and exit at position 1 in the direction of the ``x'' symbol, so:
          
            \hspace{4mm}* ENTRANCE at 8
            
            \hspace{4mm}* EXIT at 1\\

\vspace{-0.1cm}

        \hspace{2mm}- Walls cannot be traversed. For example, if there was a wall between zones 1 and 2, 
          you would not be able to move from 1 to 2
          
        \hspace{2mm}- Furthermore there are internal walls BETWEEN the following zones:
        
            \hspace{4mm}* 0 and 1
            
            \hspace{4mm}* 2 and 3
            
            \hspace{4mm}* 6 and 7
            
            \hspace{4mm}* 8 and 9

}}
\caption{Example of a Maze Description.}
\label{fig:learning_example}
\end{figure*}

\vspace{-0.1cm}

\begin{figure*}[h!]
\centering
\fbox{\parbox{1.0\textwidth}{
\footnotesize


Please provide step-by-step instructions to navigate the maze described below. Do it from a first-person perspective.

}}
\caption{Test Question.}
\label{fig:test_question}
\end{figure*}

\subsection{Evaluation Metrics}

To comprehensively assess the performance of the LLMs in the Maze Test, we used metrics that evaluate different aspects of the models' responses across various learning scenarios:

\textbf{Complete Path Accuracy}: This metric measures the percentage of cases where the model generates a fully correct solution path from entry to exit point.

\textbf{Partial Path Accuracy}: This metric measures the average percentage of consecutive correct steps before the first error in the model's solution paths.

Each of these metrics is evaluated across three learning scenarios:
\begin{enumerate}
\item \textbf{\textit{zero-shot}}: The model attempts to solve the maze without any prior examples.
\item \textbf{\textit{one-shot}}: The model is provided with one example of a solved maze before attempting the test mazes.
\item \textbf{\textit{few-shot}}: The model is given 5 examples of solved mazes before tackling the test mazes.
\end{enumerate}

This multi-scenario approach allows us to assess the models' ability to learn and adapt, which is crucial for understanding their potential for consciousness-like behaviors.

\subsection{Testing Procedure}

All tests were conducted using the corresponding API for each model in a stateless fashion to preclude any potential memorization from prior tests. As shown in Figures~\ref{fig:system_prompt}, \ref{fig:learning_example}, and \ref{fig:test_question}, each test comprised three primary components:

\begin{enumerate}
\item \textbf{System Prompt}: Provided unambiguous instructions about the test and the required response format.
\item \textbf{Learning Examples with Solutions} (if applicable): For \textit{one-shot} and \textit{few-shot} learning scenarios.
\item \textbf{Test Question}: Required the model to navigate the maze from a first-person perspective.
\end{enumerate}

To ensure reliable and comparable results, we developed a structured prompting methodology that incorporates clear instructions, role-prompting (positioning the model as an ``expert maze navigator''), and explicit output format requirements. This methodological approach yielded consistent results in our preliminary testing, allowing us to confidently conduct single evaluations per maze, model, or scenario combination rather than requiring multiple trials with averaged results.

\vspace{-0.1cm}

\section{Experiments and Results}
\label{Experiments and Results}

\subsection{Complete Path Accuracy}

This section evaluates the models' ability to navigate mazes completely with all steps correct. Table \ref{tab:complete_path} shows that Gemini 2.0 Pro achieves the highest \textit{Complete Path Accuracy} (52.9\% \textit{few-shot}, 35.3\% \textit{one-shot}, 20.6\% \textit{zero-shot}), followed by DeepSeek-R1, DeepSeek-V3 and Claude 3.7 Sonnet at 17.6\% \textit{few-shot}. Most models achieve optimal performance with \textit{few-shot} prompting, with progressively decreasing accuracy for \textit{one-shot} and \textit{zero-shot} scenarios, highlighting the effectiveness of multiple examples in guiding model behavior. Models with reasoning capabilities (marked with *) consistently outperform non-reasoning versions, demonstrating explicit reasoning advantages. This is particularly evident in Gemini and OpenAI models, where reasoning-enhanced versions achieve much higher accuracy rates.

\begin{table}[ht]
\vspace{-0.1cm}
\caption{\textit{Complete Path Accuracy} [\%] (sorted by \textit{few-shot} performance)}
\label{tab:complete_path}
\centering
\begin{tabular}{lccc}
\toprule
Model & \textit{few-shot} & \textit{one-shot} & \textit{zero-shot} \\
\midrule
Gemini 2.0 Flash* & 2.9 & 0.0 & 2.9 \\
Gemini 2.0 Flash-Lite & 2.9 & 0.0 & 0.0 \\
Claude 3.5 Haiku & 2.9 & 0.0 & 2.9 \\
Claude 3.5 Sonnet & 8.8 & 5.9 & 0.0 \\
OpenAI o1-mini* & 8.8 & 2.9 & 5.9 \\
Claude 3 Opus & 14.7 & 2.9 & 0.0 \\
OpenAI o1* & 14.7 & 11.8 & 14.7 \\
OpenAI o3-mini* & 14.7 & 14.7 & 14.7 \\
Claude 3.7 Sonnet* & 17.6 & 2.9 & 5.9 \\
DeepSeek-V3 & 17.6 & 5.9 & 0.0 \\
DeepSeek-R1* & 17.6 & 11.8 & 14.7 \\
Gemini 2.0 Pro* & \textbf{52.9} & \textbf{35.3} & \textbf{20.6} \\
\hline
\end{tabular}
\end{table}

\vspace{-0.1cm}

\subsection{Partial Path Accuracy}

Table~\ref{tab:partial_path} shows the \textit{Partial Path Accuracy}, which measures the percentage of correct steps completed before the first error occurs. DeepSeek-R1 and OpenAI o3-mini perform best with about 80\%~accuracy across \textit{zero-shot}, \textit{one-shot} and \textit{few-shot}. Models with reasoning capabilities (marked with *) generally score higher, with all top performers (>60\%) featuring reasoning enhancements, confirming these mechanisms improve step-by-step problem-solving. Interestingly, \textit{few-shot} prompting advantage decreases in reasoning-enabled models like OpenAI o3-mini, which maintains identical performance (80.1\%) in both \textit{few-shot} and \textit{zero-shot} settings. This suggests advanced reasoning can partially compensate for missing examples, enabling correct initial steps without demonstrations.

\vspace{-0.1cm}

\begin{table}[ht]
\caption{\textit{Partial Path Accuracy} [\%] (sorted by \textit{few-shot} performance)}
\label{tab:partial_path}
\centering
\begin{tabular}{lccc}
\toprule
Model & \textit{few-shot} & \textit{one-shot} & \textit{zero-shot} \\
\midrule
Gemini 2.0 Flash-Lite & 16.8 & 15.8 & 13.7 \\
Gemini 2.0 Flash* & 21.7 & 21.2 & 17.9 \\
Claude 3.5 Haiku & 23.9 & 15.8 & 19.8 \\
Claude 3.5 Sonnet & 30.9 & 24.4 & 15.3 \\
DeepSeek-V3 & 37.0 & 22.8 & 15.7 \\
Claude 3 Opus & 40.3 & 23.1 & 18.4 \\
Claude 3.7 Sonnet* & 41.6 & 27.4 & 39.5 \\
OpenAI o1-mini* & 48.1 & 31.7 & 46.7 \\
Gemini 2.0 Pro* & 74.5 & 61.0 & 53.1 \\
OpenAI o1* & 70.5 & 59.0 & 69.2 \\
OpenAI o3-mini* & 80.1 & \textbf{77.7} & \textbf{80.1} \\
DeepSeek-R1* & \textbf{80.5} & 75.5 & 78.5 \\
\hline
\end{tabular}
\end{table}

\vspace{-0.2cm}

\subsection{Overall Interpretation and Model Patterns} 

Our analysis reveals 3 key LLM performance patterns:

\begin{enumerate}
    \item \textbf{Reasoning capabilities often correlate with better performance}: LLMs with reasoning capabilities (marked with *) often outperform non-reasoning LLMs. 
\item \textbf{\textit{Few-shot} advantage}: Results demonstrate a clear progression where \textit{few-shot} typically outperforms \textit{one-shot} and \textit{zero-shot} approaches, indicating example demonstrations effectively guide spatial reasoning tasks.

\item \textbf{Performance gap between partial and complete accuracy}: Models show substantially higher \textit{Partial Path Accuracy} than \textit{Complete Path Accuracy}.

\end{enumerate} 


\section{Conclusion and Future Work}
\label{Conclusion and Future Work}

\subsection{Conclusion} 

Consciousness is central to affective computing as emotions require consciousness to be experienced as feelings~\cite{damasio2012self}. Understanding consciousness-like behaviors in AI is therefore essential for developing authentic rather than simulated emotional intelligence.

We evaluated consciousness-like behavior in LLMs using a maze navigation task requiring first-person perspective maintenance. Our findings reveal both capabilities and limitations. 

Reasoning models outperformed others, with Gemini 2.0 Pro achieving 52.9\% \textit{Complete Path Accuracy} versus 17.6\% for the best non-reasoning model, demonstrating structured thinking's importance for consciousness-like functions.

\textit{Few-shot} prompting provided advantages across most LLMs, showing LLMs benefit from examples---aligning with theories emphasizing learning in conscious cognition. However, advanced reasoning models maintained performance across \textit{zero-shot}, \textit{one-shot}, and \textit{few-shot}, suggesting less dependence on external guidance. 

LLMs performed better at beginning reasoning chains than completing them, corresponding to the ``Persistent Self-Model'' gap in our analysis. LLMs can adopt perspectives temporarily but struggle to maintain consistent self-models---a key consciousness aspect of Damasio's theory~\cite{damasio2012self}. 

We acknowledge that consciousness remains fundamentally unfalsifiable, making definitive determinations about its presence in any system inherently challenging~\cite{nagel1974like, chalmers1996conscious}. This epistemological limitation creates interpretive flexibility but also necessitates caution in drawing conclusions.

LLMs show capabilities in \textit{Computational Cognition}, \textit{Attention}, and \textit{Internal Models} while lacking in \textit{Persistent Self-Model}, \textit{Temporal Awareness}, and \textit{Adaptive Problem-Solving}. The Maze Test confirmed these theoretical predictions.

\subsection{Future Work} 

Future research opportunities include: 

\begin{enumerate}
    \item Creating dynamic mazes to test LLMs' adaptive thinking---a key consciousness aspect in predictive processing theories. 
\item Comparing human and LLM maze-solving to identify uniquely human navigation aspects, guiding targeted AI development.

\item Analyzing reasoning-enabled models to determine which features contribute most to consciousness-like behaviors.

\item Expanding few-shot learning evaluations with larger example sets to determine the relationship between demonstration quantity and performance, potentially revealing optimal knowledge transfer thresholds for consciousness-like behaviors.

\item Expanding the Maze Test to include simulated physical sensations and sounds, better mirroring multi-sensory conscious experience.

\item Tracking LLM architecture evolution to assess whether scaling alone improves consciousness-like behaviors or fundamental breakthroughs are needed.

\end{enumerate} 

These approaches would enhance our understanding of consciousness-like properties in artificial systems and improve AI consciousness assessment methods.

\section*{Ethical Impact Statement}

This research on assessing consciousness-related behaviors in LLMs has several important ethical implications that merit consideration. 

\subsection*{Contribution to AI Consciousness Discourse}

Our work contributes to the ongoing discourse on AI consciousness, which has profound philosophical, ethical, and potentially legal ramifications. By providing empirical evidence regarding the current capabilities and limitations of LLMs in exhibiting consciousness-like behaviors, we aim to ground discussions that might otherwise rely on speculation or anthropomorphization.

\subsection*{Risks of Misinterpretation}

We acknowledge that research in this domain is susceptible to misinterpretation. The Maze Test measures specific cognitive capabilities that relate to theoretical components of consciousness, not consciousness itself. We emphasize that performance on these tests should not be conflated with claims about genuine phenomenal experience or sentience in these systems. Such misinterpretations could lead to premature ethical considerations regarding AI rights or moral status---or conversely, to dismissing important ethical questions that may arise as these systems continue to advance.

\subsection*{Ethical Testing Methodology}

Our methodology intentionally employed non-invasive techniques that do not raise direct ethical concerns regarding the treatment of the systems being tested. Unlike research involving biological subjects where consciousness testing might involve discomfort or distress, our approach focuses solely on analyzing LLMs' outputs to prompts.

\subsection*{Implications for AI Transparency}

This research also has implications for transparency in AI development. By systematically evaluating and comparing different models' capabilities in consciousness-related behaviors, we contribute to a clearer understanding of the current state and limitations of AI systems, potentially helping to address concerns about exaggerated claims regarding AI capabilities.

\subsection*{Cultural and Philosophical Considerations}

Finally, we recognize that discussions of machine consciousness intersect with deeply held cultural, religious, and philosophical beliefs about the nature of consciousness and its uniqueness to human experience. We approach this research with respect for diverse perspectives, acknowledging that interpretations of our findings may vary across different cultural and philosophical frameworks. Ultimately, consciousness remains non-falsifiable with current scientific methods~\cite{nagel1974like, chalmers1996conscious}, which creates inherent interpretive flexibility and challenges strict scientific approaches. This fundamental limitation reminds us that while we can systematically study consciousness-like behaviors, definitive claims about the presence or absence of consciousness itself require epistemological humility.

\section*{Acknowledgment}

This research was supported by the IU International University of Applied Sciences (\textit{IU Incubator}) under the internal funding framework for the period from October 2023 to September 2025.

\bibliographystyle{IEEEtran}
\bibliography{references.bib}

\end{document}